\documentclass[runningheads]{llncs}
\usepackage{graphicx}
\usepackage{amssymb}
\usepackage{amsmath,tabularx}
\usepackage{xcolor}
\usepackage{multirow}
\usepackage{siunitx}
\usepackage{booktabs}
\usepackage{etoolbox}

\usepackage[ruled]{algorithm2e}
\usepackage{algorithmic}
\usepackage{todonotes}
\usepackage{bbm}

\newcommand{\printfnsymbol}[1][\value{footnote}]{\footnotemark[#1]}
\usepackage{authblk}

\begin{document}

\robustify\bfseries
        
\title{Confident Coreset for Active Learning in Medical Image Analysis}
\titlerunning{Confident Coreset for Active Learning}

\author{Seong Tae Kim\inst{1,\dagger,}\thanks{\textit{First two authors contributed equally to this work.\newline $^\dagger$ Corresponding author (seongtae.kim@tum.de)}}, Farrukh Mushtaq\inst{1,}\printfnsymbol, \and Nassir Navab\inst{1,2}}
\authorrunning{S.T. Kim et al.}
\institute{Computer Aided Medical Procedures, Technical University of Munich, Germany
\and
Computer Aided Medical Procedures, Johns Hopkins University, USA}

\maketitle              
\begin{abstract}
Recent advances in deep learning have resulted in great successes in various applications. Although semi-supervised or unsupervised learning methods have been widely investigated, the performance of deep neural networks highly depends on the annotated data. The problem is that the budget for annotation is usually limited due to the annotation time and expensive annotation cost in medical data. Active learning is one of the solutions to this problem where an active learner is designed to indicate which samples need to be annotated to effectively train a target model. In this paper, we propose a novel active learning method, confident coreset, which considers both uncertainty and distribution for effectively selecting informative samples. By comparative experiments on two medical image analysis tasks, we show that our method outperforms other active learning methods.

\end{abstract}

\section{Introduction}


Remarkable success in deep learning studies is largely attributed to the collection of large datasets with human-annotated labels. There are lots of unlabeled data available, but usually, it is extremely expensive and time-consuming to label them, both in terms of annotation labor and cost. This becomes even more apparent in medical applications where labeling needs to be done by medical experts. To address this issue, active learning approaches have been investigated to incrementally select samples for annotation which are more beneficial to the model. In active learning, we repeatedly acquire labels from an annotator only for the most informative data points from a pool of available unlabeled data within the limited budget for the annotation (pool-based active learning).

Majority of studies in pool-based active learning has been done by following two categories: 1. Uncertainty-based approaches, 2. Distribution-based approaches. The uncertainty based approaches define the quantity of uncertainty and use it as a selection criterion for labeling. Gal et al. \cite{gal2017deep} have proposed Monte-Carlo (MC) dropout as an approximate inference in Bayesian deep learning model to measure the uncertainty of samples. Beluch et al. \cite{beluch2018power} have compared the ensemble-based approaches with MC dropout. The ensemble of deep networks is implemented by multiple network architectures with different weight initialization. In the experiments, they show that MC dropout is worse than the ensemble-based method because it has reduced capacity due to the same weights initialization. Recently, Yoo et al. \cite{yoo2019learning}  have introduced a new framework where a loss prediction model is trained with a task learner at the same time. The loss prediction model is used to calculate the uncertainty of the samples in the remaining unlabeled pool. The distribution-based approaches use the diversity of the samples as a criterion to add a new sample for labeling. Sinha et al. \cite{sinha2019variational} have proposed a variational adversarial active learning method which considers distribution of labeled and unlabeled data.  
Sener et al. \cite{sener2018active} have proposed a coreset approach where the active learner finds a set of points for labeling such that the difference in performance of the model learned on the labeled dataset is minimum between the full dataset and the labeled dataset. They formulate this as \textit{K}-center problem which finds representative \textit{K}-data points to cover whole \textit{N} data points such that maximum distance for each data point to its nearest center is minimized. 

In medical community, active learning methods have been investigated for lymph node image segmentation in ultrasound and red blood cell subtype detection in microscopy \cite{yang2017suggestive,sadafi2019multiclass}. Yang et al. \cite{yang2017suggestive} statistically estimate uncertainty of samples by ensemble models. Sadafi et al. \cite{sadafi2019multiclass} selct samples for annotation by variational inference with MC dropout.

This paper introduces a novel pool-based active learning method that selects the most informative samples based on both considering uncertainty and distribution. In medical applications, a collected dataset is proportional to the distribution of patients in screening or diagnosis and it is usually class-imbalanced. Although the uncertainty-based approach has strength which can select difficult samples from the target model, it is limited when the batches of data points are acquired during each annotation step. Because uncertainty-based approaches only consider the difficulty of the model, in batch settings, they usually select the samples with high overlaps. In other words, they acquire data points that are individually informative, but not necessarily jointly informative \cite{kirsch2019batchbald}. This becomes a more serious problem in class imbalanced condition. If we reduce the number of samples for each data acquisition step to avoid this problem, the model training becomes then a computational bottleneck for larger target models and medical experts should wait until the next data acquisition time to new samples getting assigned to them \cite{kirsch2019batchbald}. On the contrary, the distribution-based approaches try to increase diversity but cannot consider difficult samples. To overcome the aforementioned limitations, we propose a new active learning model that not only increases the diversity but also considers uncertainty as an informative score for each sample. The followings are major contributions of this paper:
\begin{itemize}
    \item We devise a novel active learning model to select informative samples by considering both uncertainty and distribution at the same time: Our method is based on coreset selection problem and incorporates uncertainty from the target model into the distribution-based informative score.
    \item Comparative experiments are conducted to verify the effectiveness of the proposed method on two representative medical image analysis tasks: Experimental results show that the proposed method is effective for selecting informative samples in both classification and segmentation tasks. 
\end{itemize}

\section{Methodology}
\begin{figure*}[t]
	\centering
	\includegraphics[width=\textwidth]{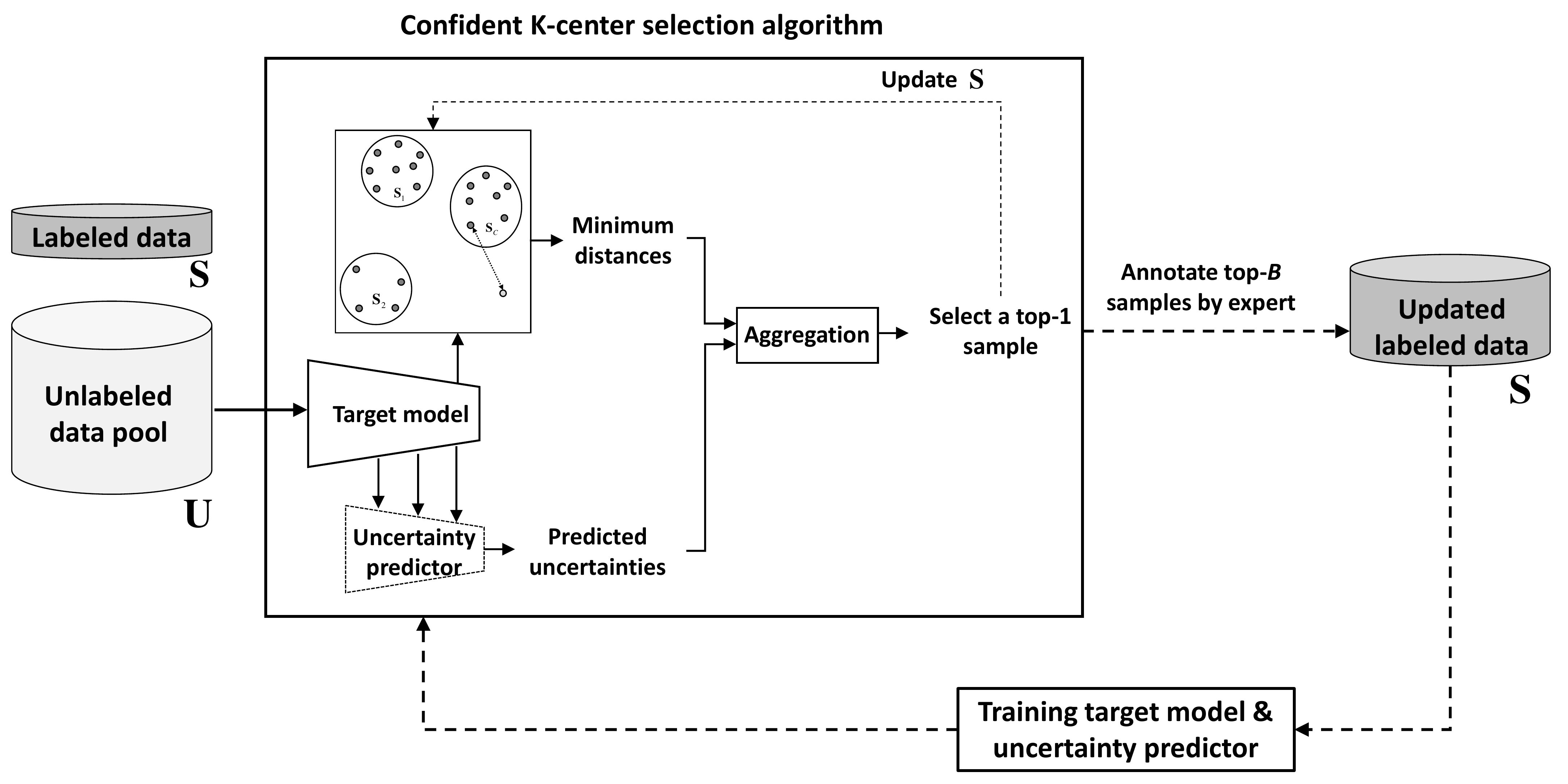}
	\caption{Overall framework of the proposed confident coreset method. A target model and an uncertainty predictor are trained from the labeled data. In active learning step, top-\textit{B} samples are selected based on confident \textit{K}-center selection algorithm.}
	\label{Figure_OverallFramework}
\end{figure*} 

\subsection{Overall Framework}
The pool-based active learning problem can be defined as followings: Let's assume that there exists a large unlabeled data pool $\mathbf{U}={\{{\mathbf{x}_{i}}\}}_{i\in [n]}$ where $[n]=\{1,\cdots ,n\}$. At the first time, a small subset of $\mathbf{U}$ is randomly selected as an initial labeled data ${{\mathbf{S}}}={{\{{\mathbf{s}}(j)\in [n]\}}_{j\in [m]}}$. 
After that, an active learning model iteratively has two stages. First, identifying a set of informative data points and presenting them to annotator to be labeled. Second, training a target model using both new and previously labeled data points in a fully supervised manner. The main problem is how to choose more informative \textit{B} samples (budget size of \textit{B}) from unlabeled data for effectively training the target model. 

The overall framework of our model is described in Fig. \ref{Figure_OverallFramework}. 
Our method is based on \cite{sener2018active} to consider distribution of labeled samples. We utilize intermediate feature space of a target model and select samples to increase diversity of labeled samples. The distribution-based approaches are more effective than the uncertainty-based approaches in real-world application scenario where batches of data points are acquired during each active learning step. Naive uncertainty-based approaches lead to acquiring data points that are individually informative, but they are not necessarily jointly informative \cite{kirsch2019batchbald}. That means there are high overlaps between selected samples. 

For given budget size (\textit{B}), our distribution-based approach repeatedly select the data point which is at a maximum distance from its closest labeled point as
\begin{equation}\label{eq-base}
\begin{aligned}
    u=\arg {{\max }_{i\in [n]\backslash \mathbf{S}}}\{\min_{j\in \mathbf{S}}\Delta ({{\mathbf{x}}_{i}},{{\mathbf{x}}_{j}}))\},
\end{aligned} 
\end{equation}
\begin{equation}\label{eq-integrage}
\begin{aligned}
    \mathbf{S}=\mathbf{S}\cup \{u\},
\end{aligned} 
\end{equation}
where $\Delta$ denotes a distance metric. 
However, it does not consider the difficult samples which can play an important role to improve the target model. 

To address this issue, we design an uncertainty predictor inspired from \cite{yoo2019learning}, which estimates the target model's uncertainty for the given sample. Note that although various approaches have been investigated to measure the uncertainty, the softmax probabilities are known to be not a good proxy of uncertainty \cite{sener2018active,yoo2019learning}. To overcome this limitation, we design an uncertainty predictor which is composed of only a few layers and jointly learned with the target model. It takes multi-layer feature maps from the target model as inputs to adaptively select meaningful information for uncertainty prediction. Each feature map is processed by global average pooling and a fully connected layer. Then, they are concatenated to produce one feature vector. Finally, one fully connected layer is applied to produce an uncertainty value. In training stage, the uncertainty predictor is trained to predict loss value of the target model for given images. In other words, it considers high loss as high uncertainty. In the original target model, it can not calculate the loss value in the test scenario because it requires knowing the true class. Instead, the uncertainty predictor is trained as a surrogate for calculating loss value at the test time. For training, a loss function for the target model $L_\text{target}$ and a loss function for the uncertainty prediction $L_\text{up}$ are used as followings
\begin{equation}
    \label{eq:loss_uncertainty}
    \frac{1}{N_b}\sum_{(\mathbf{x},y)\in\mathcal{S}_b}L_\text{target}(\hat{y},y)+\lambda\frac{2}{N_b}\cdot \sum_{(\mathbf{x}^p,y^p)\in\mathcal{S}_b}L_\text{up}(\hat{l^p}, l^p),\\
\end{equation}
where $L_\text{target}(\hat{y}, y)$ denotes a loss function to learn the target model. For classification task, cross-entropy loss is used as a target loss and for segmentation, dice loss is used as a target loss in this study.
$L_\text{up}(\hat{l^p}, l^p)$ denotes a comparative loss with margin to learn the uncertainty predictor. $\mathbf{x}^p=(\mathbf{x}_{i_1},\mathbf{x}_{i_2})$ denotes a pair of inputs where the subscript \textit{p} represents it is a pair. $\mathcal{S}_b$ denotes a mini-batch and $N_b$ is the number of samples in $\mathcal{S}_b$. $y$ is a target label and $\hat{y}$ is a target prediction. As a target uncertainty, we use a loss value of the target model. $l^p=L_\text{target}(\hat{y}, y)$ and $\hat{l^p}=G(\mathbf{x})$ denote a target loss and a predicted loss of the uncertainty predictor, respectively. $G(\mathbf{x})$ denotes a function for the uncertainty predictor that produces loss. 
To use the difference between a pair of loss predictions by avoiding the scale problem\cite{yoo2019learning}, the comparative loss with margin is used for $L_\text{up}$ as followings 
\begin{multline}
    \label{eq:uncertainty_prediction_loss}
    L_\text{up}(\hat{l^p}, l^p) = \max\left(0, -f(l_{i_1},l_{i_2})\cdot(\hat{l_{i_1}}-\hat{l_{i_2}})+\xi\right)\\
    \text{s.t.}\quad f(l_{i_1},l_{i_2})=
    \left\{\begin{aligned}
    +1,&\quad\textrm{if}\quad l_{i_1}>l_{i_2}\\
    -1,&\quad\textrm{otherwise}
    \end{aligned}\right.
\end{multline}
where $\xi$ is a pre-defined positive margin. 

\subsection{Confident \textit{K}-center Selection Algorithm}
As explained in Subsection 2.1, our model uses both the distribution and the uncertainty. Then, how to aggregate two different information is the question. The uncertainty-based score can consider "hard" samples which might be located near the decision boundary. The distribution-based score can consider the diversity of samples.
To effectively use complementary information of both uncertainty and diversity, we propose a new active learning method, confident coreset. The confident coreset merges the uncertainty and the minimum distance to calculate the informative score in the greedy K-center selection algorithm. The Eq. \ref{eq-base} can be extended as followings:
\begin{equation}
    \label{eq:Confident_coreset}
    u=\arg {{\max }_{i\in [n]\backslash \mathbf{S}}}\{(N({{\min }_{j\in \mathbf{S}}}\Delta ({{\mathbf{x}}_{i}},{{\mathbf{x}}_{j}})))^\alpha
    \times(N(G({{\mathbf{x}}_{i}})))^{1-\alpha}\},
\end{equation}        
where $N(\cdot)$ denotes the min-max normalization. $\alpha$ denotes an weighting hyperparameter to balance between the distribution-based score and the uncertainty-based score. 
The Algorithm \ref{Alg:1} shows the pseudocode of the proposed confident \textit{K}-center selection algorithm for confident coreset model in our active learning.

\begin{algorithm}[!t]
	\caption{Confident \textit{K}-center selection algorithm}

	\label{Alg:1}
	\algsetup{linenosize=\small}
	\small
	\textbf{Input:} For given data $\textbf{x}_i$, existing pool $\textbf{S}^0$ and a budget $B$
    
    Initialize $\textbf{S}=\textbf{S}^0$
    
    \For{$k=1,2,\cdots ,B$}{
        $u=\arg {{\max }_{i\in [n]\backslash \mathbf{S}}}\{(N({{\min }_{j\in \mathbf{S}}}\Delta ({{\mathbf{x}}_{i}},{{\mathbf{x}}_{j}})))^\alpha\times
        (N(G({{\mathbf{x}}_{i}})))^{1-\alpha}\}$
        
        $\mathbf{S}=\mathbf{S}\cup \{u\}$
        
        where  $N(\cdot)$ denotes the min-max normalization.
    }
    
    return $\mathbf{S}\backslash {{\mathbf{S}}^{0}}$
\end{algorithm}

\section{Experimental Setup}
\subsection{Datasets}
To evaluate the effectiveness of the proposed method, comparative experiments are conducted on two medical image analysis tasks: 1) Skin lesion classification and 2) skin lesion segmentation. For skin lesion classification, the HAM10000 dataset \cite{tschandl2018ham10000} is used. 
It contains 10,015 dermoscopic images and each image is included in one of the following seven classes: Melano-cytic nevi, melanoma, benign keratosis-like lesions, bascal cell carcinoma, actinic keratoses, vascular lesions, and dermatofibroma. 
The dataset is highly imbalanced. There are 6,705 melanocytic cases which are much high number than the dermatofibroma class where only 115 cases are included. The 80\% of the dataset is used as a training set and the remaining 20\% of the dataset is used as a test set. In the training set, 80\% is used for model training and 20\% is used for validation.   
For skin lesion segmentation task, International Skin Imaging Collaboration (ISIC) skin lesion segmentation challenge dataset (ISIC2017) \cite{codella2018skin} is used. It contains 2,000 training, 150 validation, and 600 test images. The lesion boundaries of melanoma, nevus, seborrheic keratosis are delineated as ground truth.

\subsection{Implementation Details}
As a target model, ResNet-18 \cite{he2016deep} is used for skin cancer classification and residual U-Net \cite{ronneberger2015u} (U-Net with ResNet-18 encoder) is used for skin lesion segmentation. For both models, intermediate feature maps encoded at 2,3,4,5 blocks of ReseNet-18 are utilized as an input for uncertainty prediction module. $\lambda$ in Eq. \eqref{eq:loss_uncertainty} is set to 1 and the pre-defined positive margin $\xi$ in Eq. (4) is set to 0.5. During training for both classification and segmentation, dataset samples are resized to a 224x224 size and data augmentation is used with horizontal flip. 
For the distance metric $\Delta$ in Eq. (5), the euclidean distance is used on the features obtained after applying average pooling to the last convolution layer of ResNet-18. $\alpha$ is empirically set to 0.25 in classification, and set to 0.5 in segmentation. In subsection 4.2, the effect of $\alpha$ will be discussed. 
\begin{figure*}[t]
	\centering
	\includegraphics[width=\textwidth]{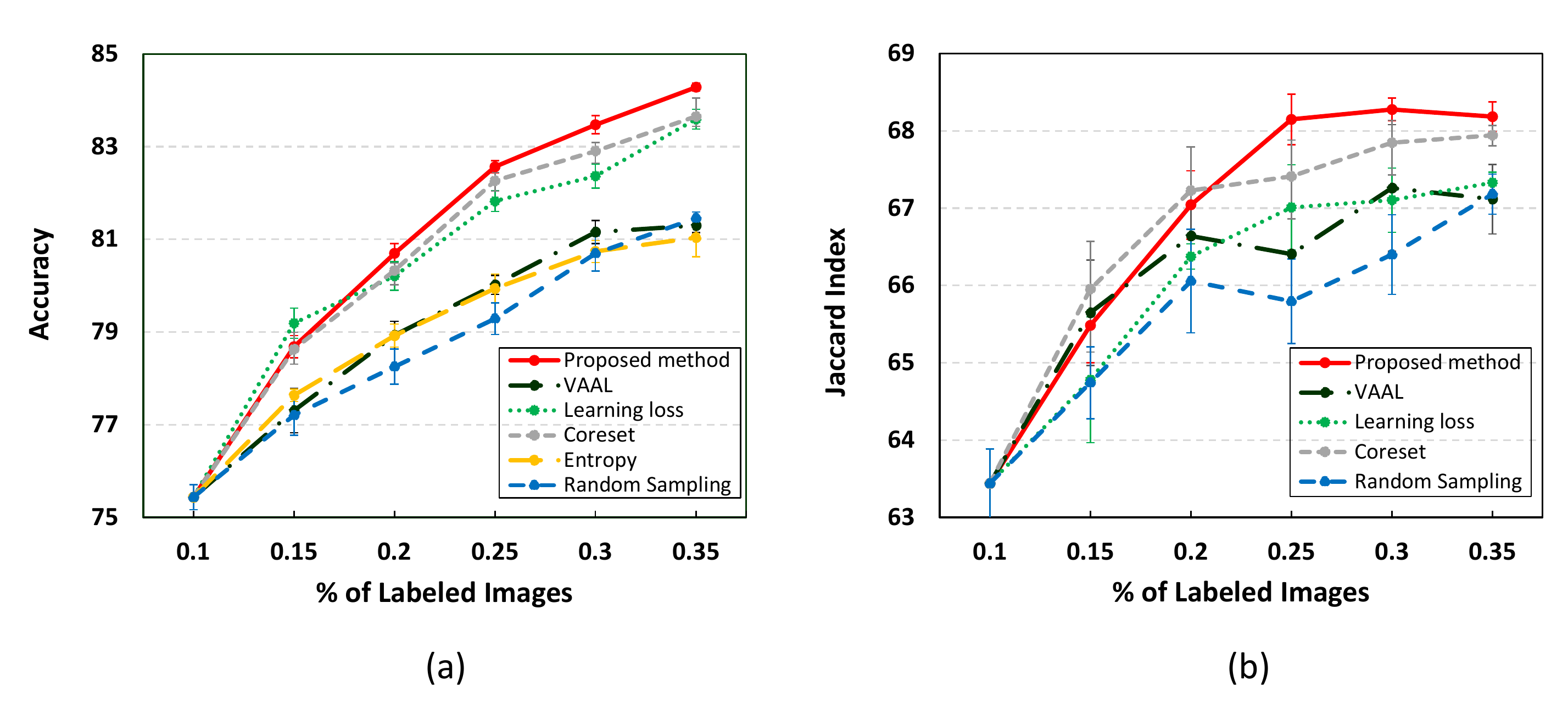}
	\caption{Comparison with other active learning methods. (a) Active learning results for classification task.  
	(b) Active learning results for segmentation task.} 
	\label{Figure_SOTA}
\end{figure*} 

\subsection{Evaluation}
For both tasks, we begin our experiments with an initial labeled data with 10\% of the training set. The budget size is set to 5\% of the full training data.  
The evaluation is done by measuring target model’s performance which is trained on the data points selected by the active learning model. We run the experiments 5 times with different initial labeled datasets to calculate mean and standard error. 
As an evaluation metric, accuracy is used for skin lesion classification and Jaccard index is used for skin lesion segmentation.

\section{Results and Discussion}
\begin{figure*}[t]
	\centering
	\includegraphics[width=\textwidth]{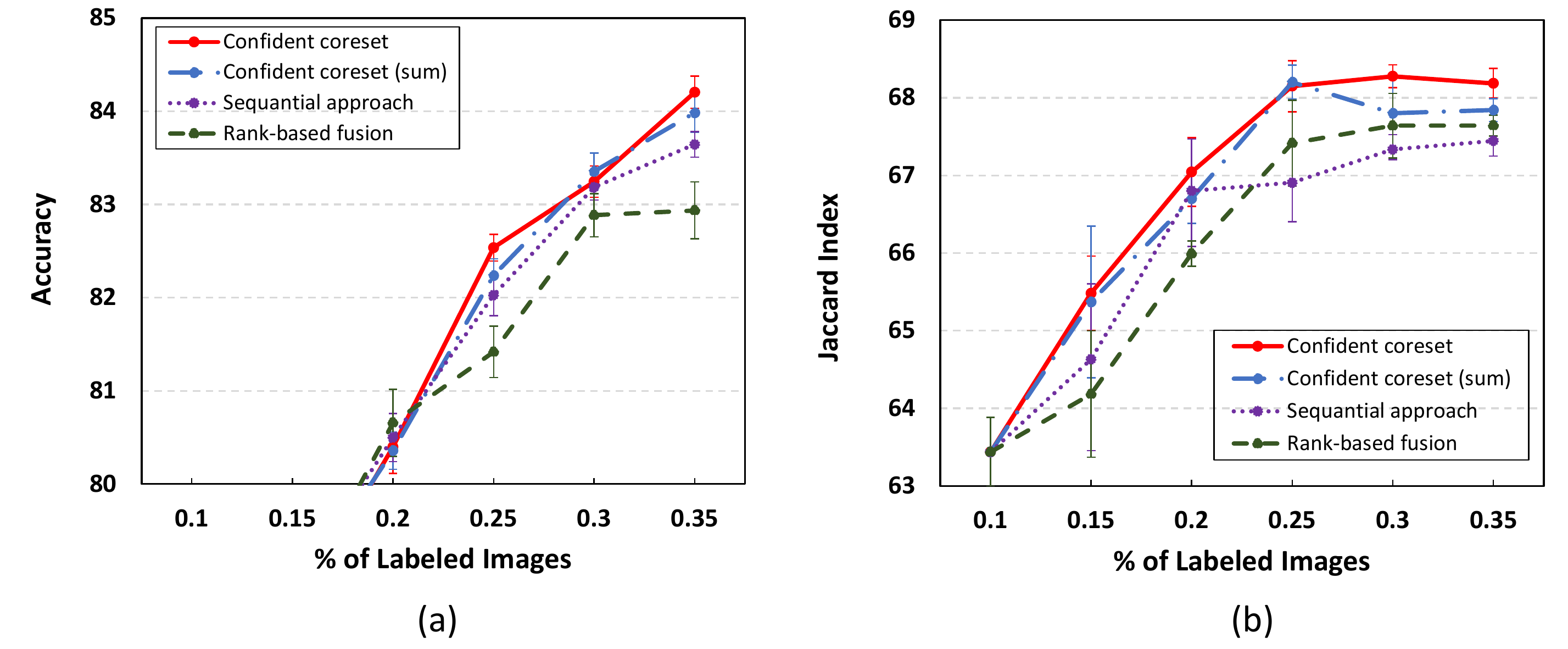}
	\caption{Comparison of different aggregation methods. (a) Active learning results for classification task. (b) Active learning results for segmentation task.}
	\label{Figure_Ablation}
\end{figure*} 
\subsection{Comparison with Other Methods}
Comparison with other methods \cite{wang2016cost,sener2018active,yoo2019learning,sinha2019variational} is conducted in this subsection. Fig. \ref{Figure_SOTA} (a) shows active learning results of skin cancer classification and (b) shows active learning results of skin lesion segmentation. We compare the proposed method with VAAL \cite{sinha2019variational}, Coreset \cite{sener2018active}, Learning loss \cite{yoo2019learning}, entropy of the softmax output \cite{wang2016cost}, and random sampling. We follow the same hyper-parameters to train the model for fair comparison. 
As shown in the figure, active learning algorithms achieve better results compared with random sampling. By looking at the number of labels to reach a fixed performance, for instance 67.04\% of Jaccard index on the segmentation task, confident coreset needs 20\% of training data (400 images) to be labeled whereas this number is approximately between 30\% (600 images) and 35\% (700 images) for random sampling.
Moreover, it is observed that coreset \cite{sener2018active} achieves higher accuracy compared with uncertainty-based methods \cite{wang2016cost,yoo2019learning}. It is mainly due to the reason that the distribution of samples is more informative compared with uncertainty in class-imbalance cases and batch setting. 
By effectively considering both uncertainty and distribution, our model could outperforms other methods. For the same target model, the maximum achievable accuracy on lesion classification is 85.2\% using 100\% of the data while our method (confident coreset) achieves 84.3\% by using 35\% of it. The maximum Jaccard incex on lesion segmentation is 69.2\% using 100\% of the data while confident coreset achieves 68.3\% by using 30\% of it.  

\subsection{Ablation Study}
\subsubsection{Aggregation strategy.} We also compare different strategies to aggregate the uncertainty-based method and the distribution-based method. Our method is compared with sequential approach and ranking-based fusion. The confident coreset (sum) is also implemented by replacing Eq. \ref{eq:Confident_coreset} with Eq. \ref{eq-summation} as
\begin{equation}\label{eq-summation}
\begin{aligned}
{u=\arg {{\max }_{i\in [n]\backslash \mathbf{S}}}\{N({{\min }_{j\in \mathbf{S}}}\Delta ({{\mathbf{x}}_{i}},{{\mathbf{x}}_{j}}))+ N(G({{\mathbf{x}}_{i}}))}\}.
\end{aligned}
\end{equation}
The rank-based fusion is implemented by assigning a score of sample based on ranks of each approach and calculating the average ranks (late fusion). The samples with high ranks are selected. The sequential approach is implemented by firstly selecting 2.5\% of unlabeled training data (half of the budget) from uncertainty-based approach and the remaining samples of half of the budget is selected by distribution-based approach. Fig. \ref{Figure_Ablation} shows the results of active learning according to the different aggregation methods. As shown in the figure, incorporating uncertainty scores in \textit{K}-center selection algorithm is better than other methods. The product is slightly better than summation for K-center selection algorithm. 
\begin{figure*}[t]
	\centering
	\includegraphics[width=\textwidth]{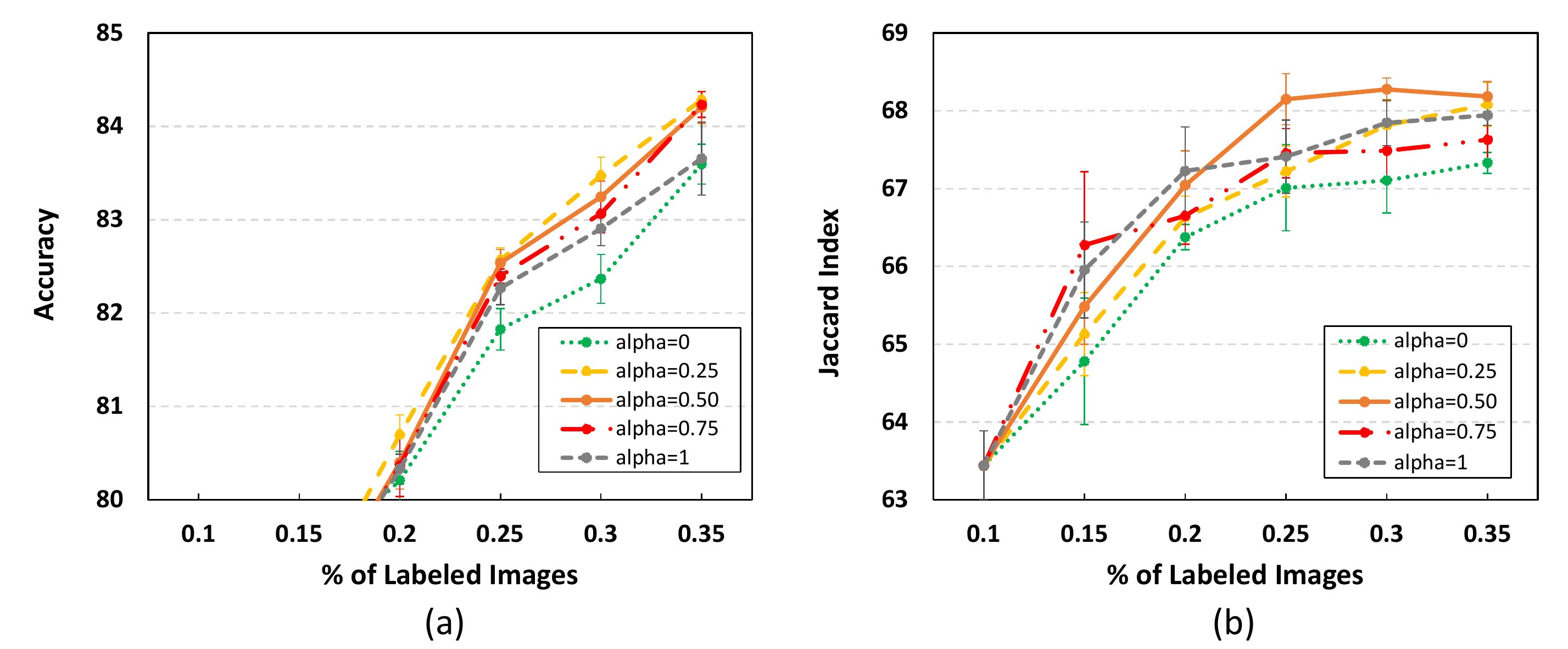}
	\caption{Analysis of effect of hyper-parameter. (a) Active learning results for classification task. (b) Active learning results for segmentation task.}
	\label{Figure_HyperParameter}
\end{figure*} 
\subsubsection{Effect of weighting parameter.} We analyze our method by changing weighting parameters $\alpha$ in [0, 0.25, 0.50. 0.75, 1]. As shown in Fig. \ref{Figure_HyperParameter}, the performances were slightly changed depending on the $\alpha$. Considering both uncertainty and distribution is better than each method ($\alpha$ of 0 or 1) over many active learning stages. 

\section{Conclusion}
In this paper, we presented a new active learning model, confident corest, for active learning in medical image analysis. To overcome the limitation of uncertainty approaches which select samples with high overlaps in batch settings, our method is based on coreset selection problem. To consider difficult samples from the view-point of the target model, we incorporate uncertainty from the target model into the distribution-based informative score. By effectively modeling the active learner to consider both the uncertainty and the distributions information, our method outperforms other active learning methods. Improving uncertainty predictor to overcome its limitation in batch active learning setting would be a meaningful agenda for future work.

\section*{Acknowledgements}
\label{sec:acknowledgements}
This study is partially supported by Google Cloud.

\bibliographystyle{splncs04}
\bibliography{references}

\end{document}